\pdfoutput=1

\documentclass[11pt]{article}

\usepackage[]{acl}

\usepackage{times}
\usepackage{latexsym}

\usepackage[T1]{fontenc}
\usepackage[utf8]{inputenc}

\usepackage{microtype}\microtypesetup{nopatch=item}

\usepackage{graphicx}
\usepackage{multirow}
\usepackage{makecell}
\usepackage{booktabs}
\usepackage{xcolor}

\newcommand{\CRAC}{CRAC 2022 Shared Task}
\newcommand{\CRAClong}{CRAC 2022 Shared Task on Multilingual Coreference Resolution}
\newcommand{\texttts}[1]{{\small\texttt{#1}}}
\newenvironment{citemize}{\begin{list}{$\bullet$}{\topsep=.1\smallskipamount\itemsep=0pt\parsep=1pt\labelwidth=.5em}}{\end{list}}
\newenvironment{cenumerate}{\begin{list}{\labelenumi}{\usecounter{enumi}\topsep=.1\smallskipamount\itemsep=0pt\parsep=1pt\labelwidth=.8em}}{\end{list}}

\title{ÚFAL~CorPipe at CRAC 2022: Effectivity of Multilingual Models for~Coreference Resolution}

\author{Milan Straka \and Jana Straková \\
  Charles University, Faculty of Mathematics and Physics \\
  Institute of Formal and Applied Linguistics \\
  Malostranské nám. 25, Prague, Czech Republic \\
  \texttt{straka,strakova@ufal.mff.cuni.cz}
  }

\usepackage[absolute]{textpos}
\begin{document}
\begin{textblock}{16}(0,0.1)\centerline{This paper was published at \textbf{CRAC 2022} -- please cite the published version {\small\url{https://aclanthology.org/2022.crac-mcr.4/}}.}\end{textblock}
\maketitle
\begin{abstract}
We describe the winning submission to the \CRAClong. Our system first solves mention detection and then coreference linking on the retrieved spans with an antecedent-maximization approach, and both tasks are fine-tuned jointly with shared Transformer weights. We report results of fine-tuning a wide range of pretrained models. The center of this contribution are fine-tuned multilingual models. We found one large multilingual model with sufficiently large encoder to increase performance on all datasets across the board, with the benefit not limited only to the underrepresented languages or groups of typologically relative languages. The source code is available at {\small\url{https://github.com/ufal/crac2022-corpipe}}.
\end{abstract}

\section{Introduction}
\label{sec:intro}

Coreference resolution is a task of identifying and clustering multiple occurrences of entities across a textual document. The \CRAClong~\cite{sharedtask-findings} features coreference resolution on 13 datasets in 10 languages, originating from the CorefUD 1.0 multilingual dataset \cite{CorefUD1.0_2021,CorefUD1.0_2022}.

Coreference resolution is often divided into two subtasks, \textit{mention detection} and \textit{coreference linking} (also \textit{clustering}). Our contribution solves these tasks neither as a purely pipeline approach with two separate sequential models, nor as an end-to-end system \cite{lee-etal-2017-end,lee-etal-2018-higher}, as is recently more common (e.g., the baseline; \citealp{pravzak2021multilingual}), but somewhere in between: We first solve mention detection and then coreference linking with an antecedent-maximization algorithm, but both tasks are jointly fine-tuned in one shared large language model. This circumvents the explosion of possible spans in an end-to-end approach, allows for a single retrieval of mentions only, while keeping the benefit of sharing the weights and training only one model for two highly related tasks (contribution \ref{contribution_architecture}).

Our architecture is a fine-tuned large language model, with experimental results leaning toward large pretrained models with better multilingual representation. We experimented with a wide range of pretrained language models \cite{devlin-etal-2019-bert,conneau-etal-2020-unsupervised,chung2021rethinking,armengol-estape-etal-2021-multilingual,RobeCzech2021,chan-etal-2020-germans,devlin-etal-2019-bert,joshi-etal-2020-spanbert,Beto2020,martin-etal-2020-camembert,huBERT2020,LitLatBERT21,mroczkowski-etal-2021-herbert,DeepPavlov2019}, of which RemBERT \cite{chung2021rethinking} proved the most effective (contribution~\ref{contribution_pretrained}).

We found \textit{multilingual models} at the center of our research attention in the \CRAC. The shared task featured datasets with sizes ranging from tiny (457 training sentences in \texttts{de} and \texttts{en~parcorfull}) to relatively large (nearly 40K training sentences in \texttts{cs~pcedt} and \texttts{cs~pdt}), all of them evaluated with equal weight (macro average). This implied that special care must be devoted to leveling the performance on all datasets. We experimented with various combinations of fine-tuned multilingual models and various sampling strategies. Although our motivation was to mitigate the poor performance on smaller specimens, we surprisingly found that one large multilingual model with sufficiently large encoder improves results on all datasets, not only the small or linguistically related ones (contribution \ref{contribution_multilinguality}).

To sum up, our contributions are the following:

\begin{cenumerate}
\item \label{contribution_architecture} We present a jointly trained pipeline approach for coreference resolution.
\item \label{contribution_pretrained} Although many monolingual base models surpass their multilingual base counterparts, in the end, one large multilingual pretrained model gives the best performance over base, albeit specifically pretrained monolingual encoders.
\item \label{contribution_multilinguality} One fine-tuned all-data multilingual model with sufficiently large encoder outperforms individual models across all datasets, not only the smaller or typologically related ones.
\end{cenumerate}
\noindent The source code of our system is available at {\small\url{https://github.com/ufal/crac2022-corpipe}}\rlap{.}

\section{Related Work}

Coreference resolution is often divided into two subtasks: \textit{mention detection} and \textit{coreference linking} (or \textit{clustering}). These can be solved either separately (pipeline approach) or, more recently, in an end-to-end fashion \cite{lee-etal-2017-end,lee-etal-2018-higher}. Such was also the approach of the baseline \cite{pravzak2021multilingual}. Our proposal takes what we hope is advantageous from both approaches: We solve both tasks sequentially, but the weights are trained jointly in a shared network.

As in all other NLP areas, deep learning with representations from large language models represents the current state-of-the-art \cite{kantor-globerson-2019-coreference,joshi-etal-2019-bert,joshi-etal-2020-spanbert}. We build on these works which use BERT \cite{devlin-etal-2019-bert} by comparing BERT with its successors, the language-specific mutations of BERT \cite{armengol-estape-etal-2021-multilingual,RobeCzech2021,chan-etal-2020-germans,devlin-etal-2019-bert,joshi-etal-2020-spanbert,Beto2020,martin-etal-2020-camembert,huBERT2020,LitLatBERT21,mroczkowski-etal-2021-herbert,DeepPavlov2019}, and multilingual variants: mBERT \cite{devlin-etal-2019-bert}, XLM-R base and XLM-R large \cite{conneau-etal-2020-unsupervised}, and RemBERT \cite{chung2021rethinking}.

There are mixed accounts in the literature on globally decoding the entities (clusters) via higher-order methods. \citet{kantor-globerson-2019-coreference} improved state-of-the-art on the CoNLL-2012 shared task with differentiable end-to-end manner enabling higher-order inference: mentions are represented as the sum of all mentions of the entity (\textit{entity equalization}). Other higher-order coreference linking methods include \textit{attended antecedent} \cite{lee-etal-2018-higher,fei-etal-2019-end,joshi-etal-2019-bert,joshi-etal-2020-spanbert}. On the other hand, \citet{xu-choi-2020-revealing} thoroughly investigated the contribution of higher-order methods to the models performance and conclude that with modern encoders, higher-order methods contribute only marginally or negatively. As we model \textit{all} antecedent links during training in a dot-product attention matrix, we inherently ``equalize'' entities, although not with an explicit algorithm.

\section{Methods}

An overview of the model architecture is shown in Figure~\ref{fig:architecture}. In the following sections, we describe the components of the model in detail, with reference to the corresponding parts of Figure~\ref{fig:architecture}.

\subsection{Architecture}

\begin{figure*}[t!]
    \centering
    \includegraphics[width=1\hsize]{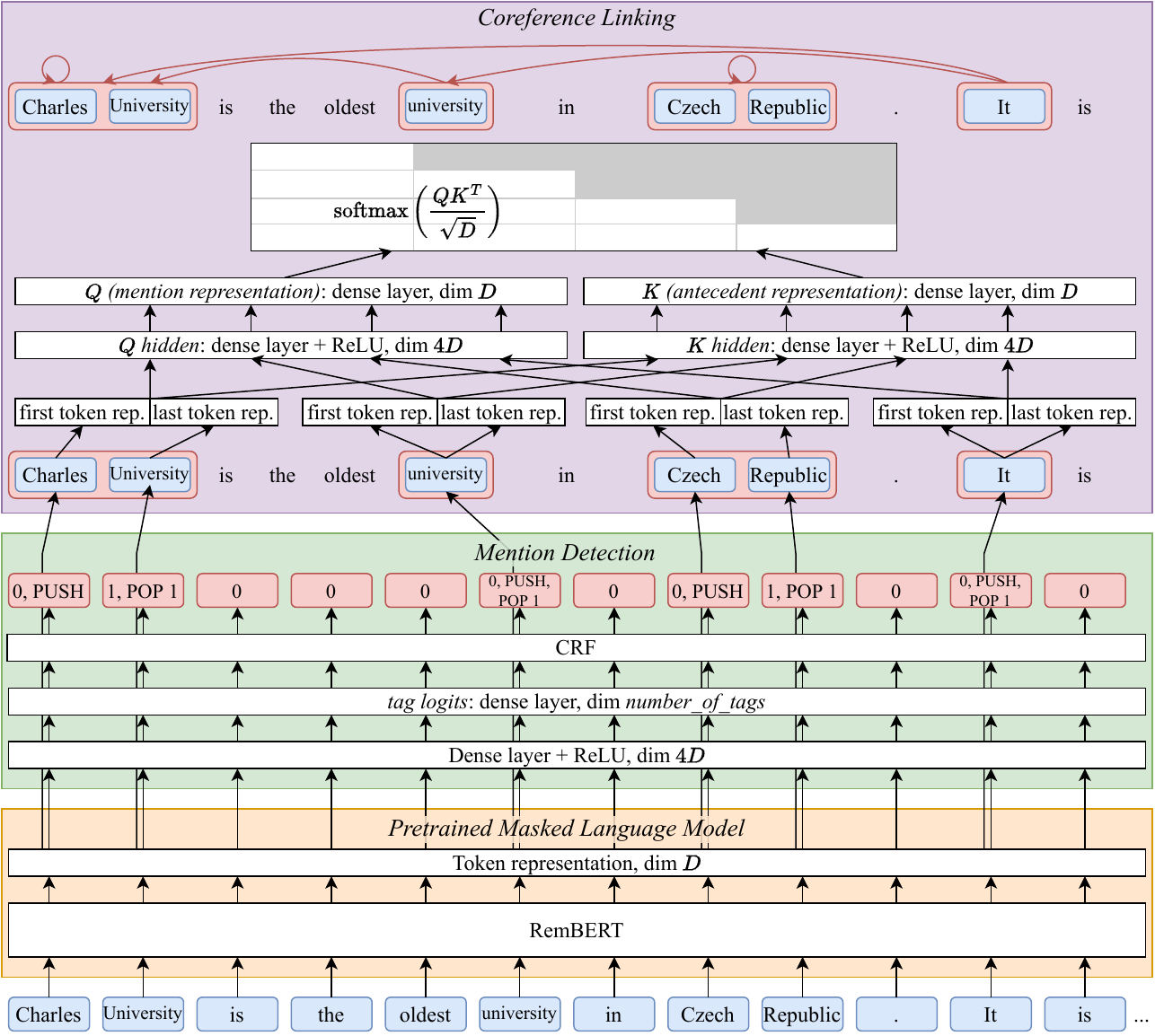}
    \caption{CorPipe model architecture. Best viewed in color.}
    \label{fig:architecture}
\end{figure*}

We consider that the enumeration of all possible spans as mention candidates in an end-to-end approach, despite aggressive pruning, may lead to explosion of options and possibly harm the coreference linking because the candidate set is heavily biased toward negative outcome: only a fraction of the spans is an actual mention and of these, only a fraction is a mention of the same entity. Furthermore, this approach does not allow the retrieval of mentions only. Hence, we propose a jointly trained, pipeline approach: we first solve \textit{mention detection} and then \textit{coreference linking} only on the retrieved mentions. However, to share the information between these highly related tasks and to keep a single model, we fine-tune one shared large language model, only with separately stacked hidden layers on top of the shared large language model for each task. In Figure~\ref{fig:architecture}, the orange box corresponds to the shared fine-tuned large language model (encoder), the green box corresponds to the \textit{mention detection} task and the purple box corresponds to the \textit{coreference linking} task.

In all fairness it should be said that we did not experimentally compare our architecture with the purely pipeline models with separate encoders nor the end-to-end approach, as pursuing three architectures to final submission was beyond our means in the given time frame. We venture to suggest that separately trained pipeline models might have the advantage of greater capacity (separate large language model for each task) but might become expensive as both models must be separately fine-tuned and hyperparameter-searched.

\subsection{Token Representations (Encoder)}

Each token receives a contextualized representation from the encoder, a vector of dimension $D$ ($D=768$ for base encoders, $D=1024$ for XLM-R large, $D=1152$ for RemBERT). The retrieval of contextualized token representation corresponds to the orange box in the bottom of Figure~\ref{fig:architecture}. The representation is shared between the \textit{mention detection} and \textit{coreference linking} tasks.

We experimented with the following pretrained multilingual language models: \begin{citemize}
\item RemBERT \cite{chung2021rethinking},
\item XLM-R base and large \cite{conneau-etal-2020-unsupervised},
\item mBERT \cite{devlin-etal-2019-bert},
\end{citemize}

\noindent and the following published language-specific models:\footnote{For each language, we present the monolingual model that worked best in our settings, if more exist.}

\begin{citemize}
\item Catalan BERTa \cite{armengol-estape-etal-2021-multilingual},
\item Czech RoBERTa RobeCzech \cite{RobeCzech2021},
\item German gBERT \cite{chan-etal-2020-germans},
\item English SpanBERT \cite{joshi-etal-2020-spanbert},
\item Spanish BETO \cite{Beto2020},
\item French CamemBERT \cite{martin-etal-2020-camembert},
\item Hungarian HuBERT \cite{huBERT2020},
\item Lithuanian LitLatBERT \cite{LitLatBERT21},
\item Polish HerBERT \cite{mroczkowski-etal-2021-herbert},
\item Russian RuBERT \cite{DeepPavlov2019}.
\end{citemize}

\subsection{Empty Nodes}

Some dependency grammar annotation schools allow, or even require, the so-called \textit{empty nodes}, which are superficial nodes of a dependency graph unseen on the surface level, i.e., not directly corresponding to any surface token of the sentence. The empty nodes usually account for ellipsis, such as in a sentence \textit{``Mary likes roses and John (likes) violets.''}, in which the verb ``likes'' is omitted but depending on the annotation guidelines may be reconstructed in the dependency tree. These nodes may carry coreference annotation, and they very often do in pro-drop languages (like Slavic or Romance languages). In Czech example: \textit{''Řekl, že nepřijde.''}, translated as \textit{''(He) said that (he) won't come.''}, both pronouns are dropped but implied by the morphology of the verb.

To allow the empty nodes, if occurring, to be naturally represented on the input and the output of the fine-tuned model and be part of the fine-tuning, we simply draw them to the surface, that is, we create a new token occupying the implied position of the artificial empty node and assign whatever text that was annotated with it (or none).\footnote{We use the form associated with a given empty node; if empty, we fall back to the (possibly empty) lemma.} To recognize such artificial tokens from regular tokens, we prepend an artificial special character to any such token originating from an empty-node.

\subsection{Mention Detection}

We model mention detection as a sequence token-level classification problem, which considers a sequence of tokens on the input and a corresponding sequence of tags on the output. The proposed tags are an extension of BIO encoding, which in addition can handle embedded and also overlapping mention spans. Each tag is a sequence of the following stack manipulation instructions:
\begin{citemize}
  \item $0..N$ \texttts{POP} instructions, each closing a mention from the stack. To handle crossing mention spans, the instruction has a parameter specifying which mention to close using its index from the top of the stack. The most frequently used value is $1$ (the top of the stack), because closing the mention on the top of the stack is sufficient to encode arbitrarily embedded non-crossing mention spans.
  \item $0..N$ \texttts{PUSH} instructions, each starting a new mention on the top of the stack.
  \item $0..N$ \texttts{POP} instructions again, each closing a single-token mention started by a previous \texttts{PUSH} in the same step. We could represent such single-word mentions using specialized \texttts{UNIT} instructions instead of a \texttts{PUSH}-\texttts{POP} pair, but we opted for less instructions for the simplicity of the decoder.
\end{citemize}
\noindent The above mentioned stack instructions are concatenated into a single tag, predicted by a classifier as one label per token.

Because not all sequences of   tags are valid (i.e., we are performing structured prediction), we process the tags by a linear-chain CRF. Finally, in order to allow the CRF to check whether there is a mention to be closed by a \texttts{POP} instruction, we include the size of the stack in the tag.\footnote{Our approach does not handle discontinuous mentions. While we could support them by introducing an instruction continuing an already closed span, handling discontinuous mentions would also require support in the mention encoder.}

The mention detection classifier corresponds to the green box in Figure~\ref{fig:architecture}. Token representation of dimension $D$ is processed by a hidden ReLU layer of dimension $4D$, then by a linear layer producing tag logits, and finally by a CRF layer.

\subsection{Coreference Linking}

We approach coreference linking by considering, for each mention, a probability distribution of the preceding mentions in the previous context (more on context window in Section~\ref{sec:methods-context}) being antecedents of the current mention. We also include the mention itself in the distribution, and consider it a technical antecedent if the mention has no antecedents.

During training, our goal is to predict \textit{all} mention antecedents using a categorical cross-entropy loss. During prediction, however, we predict only the most probable antecedent for every mention, noting that any correct antecedent results in the same coreference cluster.\footnote{This is true only when considering previous mentions as antecedents; if we considered both previous and following mentions as antecedents, disconnected components of a single coreference cluster could be formed.}

The computation of the antecedent distribution, corresponding to the purple box in Figure~\ref{fig:architecture}, starts by constructing an initial representation of every mention by concatenating the token representations of its first and last tokens.\footnote{Such an approach assumes the mentions are continuous. We handle discontinuous mentions by limiting them to their largest continuous sub-span containing the syntactic head of the mention (see Section~\ref{sec:limit_spans_to_heads}).} Using this representation, we compute $Q$ (the representation of a reference candidate) and $K$ (the representation of an antecedent candidate), both using a hidden ReLU layer with dimensionality $4D$ followed by a bias-free linear layer of dimensionality $D$. Finally, we compute the antecedent distribution using masked dot-product self-attention~\cite{vaswani-etal-2017-attention}.

The inclusion of ``self'' in the pool of antecedents naturally allows for the so-called \textit{singletons}, which are mentions without antecedent (entities mentioned only once, for example ``Czech Republic'' in Figure~\ref{fig:architecture}). Singletons were excluded from the official evaluation primary metric, but the official evaluation with singletons on the test set and the ablation experiments with singletons on the dev set can be found in Section~\ref{sec:results-singletons}.

The fact that during training a reference should recognize all its antecedents might seem inconsistent with the inference regime, where only a single most probable antecedent is retrieved. We therefore demonstrate the effectiveness of considering all antecedents by also evaluating a strategy of limiting the number of gold antecedents to at most 1, 2, or 3 previous ones (\textbf{\itshape{At most 1 link}}, \textbf{\itshape{At most 2 links}} or \textbf{\itshape{At most 3 links}}) in Section~\ref{sec:results-links}.

\subsection{Context Window}
\label{sec:methods-context}

For each sentence, we consider a sliding context of 512 tokens, aligning the end of the current sentence towards the end of the window to allow for a larger left (past) context than the right (future) context. We experiment with several settings of the size of the right context in Section~\ref{sec:results-right_context}:

\begin{citemize}
    \item \textbf{\itshape{Right context 0}}: The end of the current sentence is perfectly aligned with the context of 512 tokens (no right context).
    \item \textbf{\itshape{Right context 50}}: We leave 50 tokens for the right (future) context after the sentence end and whatever remains is the left (past) context; if there is not enough left context to fill the whole window of 512 tokens (e.g., the first sentence of the document), we increase the size of the right context to fill all the 512 tokens.
    \item \textbf{\itshape{Right context 100}}: Same as before, but 100 tokens for the right context.
\end{citemize}
\noindent Unless stated otherwise, we use right context of 50.

\subsection{Multilingual Models}

\begin{figure}
    \centering
    \includegraphics[width=\hsize]{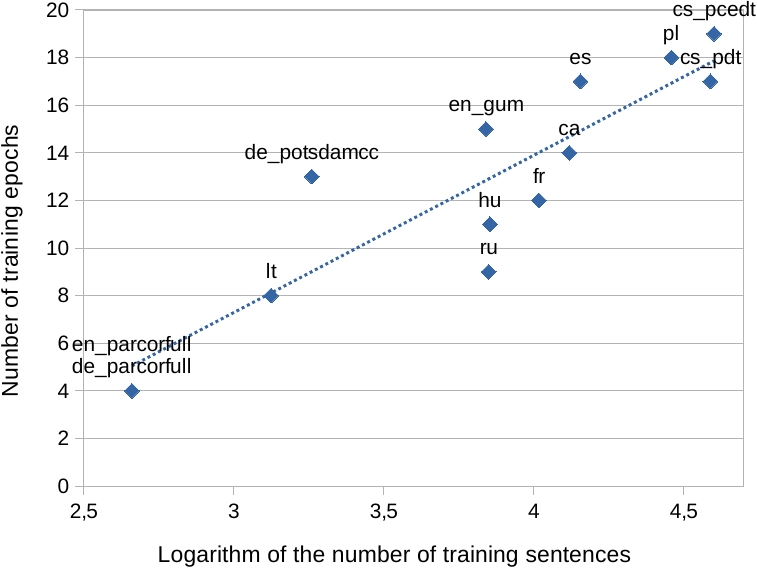}
    \caption{Dependency of the number of the optimum training epochs on the logarithm of the corpus size.}
    \label{fig:training_length}
\end{figure}

We introduced \textit{multilinguality} as our natural research interest of \CRAC. We experimented with various combinations of models with respect to size and/or language, and in the end, we submitted three contributions to the final evaluation:

\begin{citemize}
    \item \textbf{\itshape{individual}}: The models were fine-tuned using solely the training data of the corresponding dataset.\footnote{With the exception of \texttt{de} and \texttt{en~parcorfull} -- these corpora are extremely small (457 sentences each) and translations of each other, so we always train on a concatenation of them when finetuning an individual language model.}
    \item \textbf{\itshape{multilingual}}: All training data were used for fine-tuning a single multilingual model, with examples sampled according to the logarithm of the individual dataset sizes. The final checkpoint of the last training epoch was used for prediction, so there is only one single large model for all datasets, an option which could most easily qualify as a deliverable software product for multilingual coreference resolution.
    \item \textbf{\itshape{best dev}}: In this setting we considered, for every dataset, the test set prediction corresponding to a model and optimum epoch achieving the best development set performance. An intuition behind this decision is that in the multilingual settings, the smaller datasets converge sooner, while the large ones need more iterations. This is supported by the seemingly linear relationship between the logarithm of the number of training sentences and the optimum number of training epochs in Figure~\ref{fig:training_length}.

\end{citemize}

When mixing multilingual data, sampling ratios of the datasets must be determined. We experimented with three strategies for sampling the datasets; the examples are then always sampled uniformly from the chosen dataset:

\begin{citemize}
    \item \textbf{\itshape{logarithmic}}: Datasets are sampled with the probability reflecting the logarithm of their size.\footnote{We scaled the logarithmic sampling rations to the range 1~to~5, and rounded them for convenience.}
    \item \textbf{\itshape{uniform}}: Datasets are sampled with uniform probability.
    \item \textbf{\itshape{linear}}: Datasets are sampled in proportion linear to their size, which effectively equals to sampling the examples uniformly from the concatenation of the datasets.
\end{citemize}

Finally, dataset labels (corpus ids) may or may not be be added to the input to discriminate the origins. We call these settings \textbf{\itshape{w/ corpus id}} and \textbf{\itshape{w/o corpus id}}.

We compare all the above mentioned strategies for creating multilingual models in Section~\ref{sec:results-multilingual}.

\subsection{Limiting Mention Spans to Their Heads}
\label{sec:limit_spans_to_heads}

The official \CRAC~evaluation relied on the lenient \textit{partial matching}, which considers mention span correctly detected if it contains the syntactic head of the gold mention and at the same time, the predicted mention span does not include any tokens outside the gold mention span. Hence it seems prudent to not ``overpredict'' too long mention spans and prune the predicted mention spans to their syntactic head, given that syntactic analysis is available in the data. We show ablation results including the full mention spans in Section~\ref{sec:results-heads}.

\subsection{Training}

We trained our models using a lazy variant of the Adam optimizer \cite{kingma-and-ba-2015}, with a batch size of 8. The \textit{base} variants were fine-tuned on a single 16GB GeForce/Quadro GPU, using a slanted triangular learning rate schedule -- first linearly increasing from 0 to $2\cdot10^{-5}$ in the first 10\% of the training, and then linearly decaying to 0 at the end of the training. The multilingual models were trained for 30 epochs, each consisting of 6000 batches; the individual models were trained for up to 100 epochs depending on dataset size.

The \textit{large} models required fine-tuning on two 25GB GeForce GPUs, the peak learning rate was $10^{-5}$ , the multilingual models were trained for 20 epochs and the individual models up to 50 epochs. We trained 8 large multilingual models (each taking 42 hours), considering both XLM-R large and RemBERT, uniform and logarithmic mixing, presence of corpus id, and $\beta_2=0.99$ in addition to the default one. The best-performing model uses RemBERT, logarithmic mixing without corpus id, and default $\beta_2$.

\section{Results}
\label{section:results}

\begin{table*}[t]
    \centering
    \setlength{\tabcolsep}{3pt}\small
    \catcode`@ = 13\def@{\bfseries}
    \begin{tabular}{lcccccccccccccc}
      \toprule
        Team/Submission & Avg &
        \texttt{ca} &
        \makecell[c]{\texttt{cs} \\ \texttt{pcedt}} &
        \makecell[c]{\texttt{cs} \\ \texttt{pdt}} &
        \makecell[c]{\texttt{de} \\ \texttt{parc}} &
        \makecell[c]{\texttt{de} \\ \texttt{pots}} &        
        \makecell[c]{\texttt{en} \\ \texttt{gum}} &
        \makecell[c]{\texttt{en} \\ \texttt{parc}} &
        \texttt{es} &
        \texttt{fr} &
        \texttt{hu} &
        \texttt{lt} &
        \texttt{pl} &
        \texttt{ru} \\
    \midrule
      \makecell[l]{\textbf{ÚFAL CorPipe} \\ \quad\textbf{\itshape{best dev}}} & @\makecell[c]{70.72 \\ 1} & \makecell[c]{78.18 \\ 2} & @\makecell[c]{78.59 \\ 1} & @\makecell[c]{77.69 \\ 1} & \makecell[c]{65.52 \\ 2} & \makecell[c]{70.69 \\ 2} & \makecell[c]{72.50 \\ 2} & @\makecell[c]{39.00 \\ 1} & @\makecell[c]{81.39 \\ 1} & @\makecell[c]{65.27 \\ 1} & \makecell[c]{63.15 \\ 3} & @\makecell[c]{69.92 \\ 1} & \makecell[c]{78.12 \\ 2} & @\makecell[c]{79.34 \\ 1} \\
      \makecell[l]{\textbf{ÚFAL CorPipe} \\ \quad\textbf{\itshape{multilingual}}} & \makecell[c]{69.56 \\ 2} & @\makecell[c]{78.49 \\ 1} & \makecell[c]{78.49 \\ 2} & \makecell[c]{77.57 \\ 2} & \makecell[c]{59.94 \\ 3} & @\makecell[c]{71.11 \\ 1} & @\makecell[c]{73.20 \\ 1} & \makecell[c]{33.55 \\ 3} & \makecell[c]{80.80 \\ 2} & \makecell[c]{64.35 \\ 3} & \makecell[c]{63.38 \\ 2} & \makecell[c]{67.38 \\ 3} & @\makecell[c]{78.32 \\ 1} & \makecell[c]{77.74 \\ 2} \\
      \makecell[l]{UWB \\
      \quad\itshape{ondfa}$^\dagger$}                            & \makecell[c]{67.64 \\ 3} & \makecell[c]{70.55 \\ 4} & \makecell[c]{74.07 \\ 4} & \makecell[c]{72.42 \\ 4} & @\makecell[c]{73.90 \\ 1} & \makecell[c]{68.68 \\ 3} & \makecell[c]{68.31 \\ 4} & \makecell[c]{31.90 \\ 4} & \makecell[c]{72.32 \\ 4} & \makecell[c]{61.39 \\ 4} & @\makecell[c]{65.01 \\ 1} & \makecell[c]{68.05 \\ 2} & \makecell[c]{75.20 \\ 4} & \makecell[c]{77.50 \\ 3} \\
      \makecell[l]{\textbf{ÚFAL CorPipe} \\ \quad\textbf{\itshape{individual}}} & \makecell[c]{64.30 \\ 4} & \makecell[c]{76.34 \\ 3} & \makecell[c]{77.87 \\ 3} & \makecell[c]{76.76 \\ 3} & \makecell[c]{36.50 \\ 5} & \makecell[c]{56.65 \\ 5} & \makecell[c]{70.66 \\ 3} & \makecell[c]{23.48 \\ 5} & \makecell[c]{78.78 \\ 3} & \makecell[c]{64.94 \\ 2} & \makecell[c]{62.94 \\ 4} & \makecell[c]{61.32 \\ 6} & \makecell[c]{73.36 \\ 5} & \makecell[c]{76.26 \\ 4} \\
      \makecell[l]{Barbora Dohnalová \\
      \quad\itshape{berulasek}}                        & \makecell[c]{59.72 \\ 5} & \makecell[c]{64.67 \\ 5} & \makecell[c]{70.56 \\ 5} & \makecell[c]{67.95 \\ 5} & \makecell[c]{38.50 \\ 4} & \makecell[c]{57.70 \\ 4} & \makecell[c]{63.07 \\ 5} & \makecell[c]{36.44 \\ 2} & \makecell[c]{66.61 \\ 5} & \makecell[c]{56.04 \\ 5} & \makecell[c]{55.02 \\ 5} & \makecell[c]{65.67 \\ 4} & \makecell[c]{65.99 \\ 6} & \makecell[c]{68.17 \\ 5} \\
      \makecell[l]{UWB \\
      \quad\textsc{BASELINE}$^\ddagger$}                            & \makecell[c]{58.53 \\ 6} & \makecell[c]{63.74 \\ 6} & \makecell[c]{70.00 \\ 6} & \makecell[c]{67.27 \\ 6} & \makecell[c]{33.75 \\ 6} & \makecell[c]{55.44 \\ 6} & \makecell[c]{62.59 \\ 6} & \makecell[c]{36.44 \\ 2} & \makecell[c]{65.98 \\ 6} & \makecell[c]{55.55 \\ 6} & \makecell[c]{52.35 \\ 6} & \makecell[c]{64.81 \\ 5} & \makecell[c]{65.34 \\ 7} & \makecell[c]{67.66 \\ 6} \\
      \makecell[l]{Matouš Moravec \\
      \quad\itshape{moravec}}          & \makecell[c]{55.05 \\ 7} & \makecell[c]{58.25 \\ 7} & \makecell[c]{68.19 \\ 7} & \makecell[c]{64.71 \\ 7} & \makecell[c]{31.86 \\ 7} & \makecell[c]{52.84 \\ 7} & \makecell[c]{59.15 \\ 7} & \makecell[c]{36.44 \\ 2} & \makecell[c]{62.01 \\ 7} & \makecell[c]{54.87 \\ 7} & \makecell[c]{52.00 \\ 7} & \makecell[c]{59.49 \\ 7} & \makecell[c]{63.40 \\ 8} & \makecell[c]{52.49 \\ 7} \\
    \bottomrule
        \end{tabular}
    \caption{Official results of \CRAC~on the test set (CoNLL score in \%). The systems $^\dagger$ and $^\ddagger$ are described in \citet{sharedtask-prazak} and \citet{pravzak2021multilingual}, respectively; the rest in \citet{sharedtask-findings}.}
    \label{tab:official_results}
\end{table*}

\begin{table*}[t]
    \centering
    \setlength{\tabcolsep}{3.3pt}\small
    \catcode`@ = 13\def@{\bfseries}
    \begin{tabular}{lrrrrrrrrrrrrrr}
      \toprule
        Experiment & Avg &
        \texttt{ca} &
        \makecell[c]{\texttt{cs} \\ \texttt{pcedt}} &
        \makecell[c]{\texttt{cs} \\ \texttt{pdt}} &
        \makecell[c]{\texttt{de} \\ \texttt{parc}} &
        \makecell[c]{\texttt{de} \\ \texttt{pots}} &
        \makecell[c]{\texttt{en} \\ \texttt{gum}} &
        \makecell[c]{\texttt{en} \\ \texttt{parc}} &
        \texttt{es} &
        \texttt{fr} &
        \texttt{hu} &
        \texttt{lt} &
        \texttt{pl} &
        \texttt{ru} \\
      \midrule
XLM-R base, multilingual & \textcolor{black}{67.8} & \textcolor{black}{77.1} & \textcolor{black}{75.8} & \textcolor{black}{74.3} & \textcolor{black}{54.7} & \textcolor{black}{66.9} & \textcolor{black}{70.1} & \textcolor{black}{38.5} & \textcolor{black}{77.6} & \textcolor{black}{64.2} & \textcolor{black}{62.3} & \textcolor{black}{69.4} & \textcolor{black}{73.3} & \textcolor{black}{76.6} \\
Best base model, individual & \textcolor{red!100.0!black}{-5.2} & \textcolor{red!100.0!black}{-4.0} & \textcolor{blue!54.7!black}{+1.5} & \textcolor{blue!66.0!black}{+2.2} & \textcolor{red!100.0!black}{-18.2} & \textcolor{red!95.1!black}{-9.8} & \textcolor{red!100.0!black}{-3.4} & \textcolor{red!100.0!black}{-15.0} & \textcolor{red!100.0!black}{-2.4} & \textcolor{red!100.0!black}{-2.4} & \textcolor{red!100.0!black}{-2.0} & \textcolor{red!100.0!black}{-8.1} & \textcolor{blue!0.2!black}{+0.0} & \textcolor{red!100.0!black}{-5.6} \\
Best base model, best dev & \textcolor{blue!13.6!black}{+0.4} & \textcolor{red!13.8!black}{-0.6} & \textcolor{blue!54.7!black}{+1.5} & \textcolor{blue!66.0!black}{+2.2} & \textcolor{blue!18.6!black}{+2.0} & \textcolor{blue!14.3!black}{+0.6} & \textcolor{red!30.5!black}{-1.0} & \textcolor{red!6.1!black}{-0.9} & \textcolor{blue!33.2!black}{+1.2} & \textcolor{red!45.0!black}{-1.1} & \textcolor{blue!57.3!black}{+0.6} & \textcolor{blue!64.8!black}{+0.4} & \textcolor{red!0.0!black}{+0.0} & \textcolor{blue!9.3!black}{+0.2} \\
RemBERT, multilingual & \textcolor{blue!60.9!black}{+1.8} & \textcolor{blue!100.0!black}{@+1.4} & \textcolor{blue!96.4!black}{+2.6} & \textcolor{blue!96.4!black}{+3.3} & \textcolor{blue!48.3!black}{+5.2} & \textcolor{blue!100.0!black}{@+4.2} & \textcolor{blue!100.0!black}{@+3.1} & \textcolor{red!32.8!black}{-4.9} & \textcolor{blue!84.3!black}{+3.2} & \textcolor{blue!12.4!black}{+0.1} & \textcolor{blue!100.0!black}{@+1.1} & \textcolor{red!24.8!black}{-2.0} & \textcolor{blue!100.0!black}{@+5.0} & \textcolor{blue!40.7!black}{+1.1} \\
RemBERT, individual & \textcolor{red!67.1!black}{-3.5} & \textcolor{red!17.8!black}{-0.7} & \textcolor{blue!73.7!black}{+2.0} & \textcolor{blue!72.5!black}{+2.5} & \textcolor{red!100.0!black}{-18.2} & \textcolor{red!100.0!black}{-10.3} & \textcolor{blue!18.8!black}{+0.6} & \textcolor{red!100.0!black}{-15.0} & \textcolor{blue!30.6!black}{+1.2} & \textcolor{blue!68.6!black}{+0.7} & \textcolor{blue!60.0!black}{+0.7} & \textcolor{red!100.0!black}{-8.1} & \textcolor{blue!0.2!black}{+0.0} & \textcolor{red!6.8!black}{-0.4} \\
RemBERT, best dev & \textcolor{blue!100.0!black}{@+3.0} & \textcolor{blue!78.3!black}{+1.1} & \textcolor{blue!100.0!black}{@+2.7} & \textcolor{blue!100.0!black}{@+3.4} & \textcolor{blue!100.0!black}{@+10.8} & \textcolor{blue!90.0!black}{+3.8} & \textcolor{blue!77.6!black}{+2.4} & \textcolor{blue!100.0!black}{@+0.5} & \textcolor{blue!100.0!black}{@+3.8} & \textcolor{blue!100.0!black}{@+1.0} & \textcolor{blue!79.1!black}{+0.9} & \textcolor{blue!100.0!black}{@+0.5} & \textcolor{blue!96.0!black}{+4.8} & \textcolor{blue!100.0!black}{@+2.7} \\
      \bottomrule
    \end{tabular}
    \caption{Official results of ablation experiments on the test set (CoNLL score in \%).}
    \label{tab:official_ablations}
\end{table*}

\begin{table}
    \centering
    \small
    \begin{tabular}{lc}
        \toprule
         Team/Submission & Avg. with singletons \\
         \midrule
         \textbf{ÚFAL CorPipe, \textit{best dev}} & \textbf{72.98} \\
         \textbf{ÚFAL CorPipe, \textit{multilingual}} & 71.81 \\
         \textbf{ÚFAL CorPipe, \textit{individual}} & 67.93 \\
         UWB, \textit{ondfa} & 58.06 \\
         Barbora Dohnalová, \textit{berulasek} & 50.84 \\
         UWB, \textsc{BASELINE} & 49.69 \\
         Matouš Moravec, \textit{moravec} & 46.79 \\
         \bottomrule
    \end{tabular}
    \caption{Official results of evaluation with singletons on~the test set.}
    \label{tab:official_singletons}
\end{table}

\begin{table*}[p]
    \centering
    \setlength{\tabcolsep}{2.75pt}\small
    \renewcommand*{\arraystretch}{0.945}
    \catcode`@ = 13\def@{\bfseries}
    \begin{tabular}{lrrrrrrrrrrrrrr}
      \toprule
        Experiment & Avg &
        \texttt{ca} &
        \makecell[c]{\texttt{cs} \\ \texttt{pcedt}} &
        \makecell[c]{\texttt{cs} \\ \texttt{pdt}} &
        \makecell[c]{\texttt{de} \\ \texttt{parc}} &
        \makecell[c]{\texttt{de} \\ \texttt{pots}} &
        \makecell[c]{\texttt{en} \\ \texttt{gum}} &
        \makecell[c]{\texttt{en} \\ \texttt{parc}} &
        \texttt{es} &
        \texttt{fr} &
        \texttt{hu} &
        \texttt{lt} &
        \texttt{pl} &
        \texttt{ru} \\
      \midrule
      \multicolumn{15}{c}{\textsc{A) The Effect of Using Full Mentions Instead of Only Their Heads}} \\
CorPipe multilingual & 73.2 & 76.9 & 79.3 & 78.1 & 70.6 & 74.7 & 74.8 & 61.2 & 80.9 & 67.4 & 64.6 & 76.0 & 75.2 & 71.4 \\
+ full mentions & \textcolor{red!100.0!black}{-1.8} & \textcolor{red!100.0!black}{-2.4} & \textcolor{red!100.0!black}{-1.2} & \textcolor{red!100.0!black}{-0.9} & \textcolor{red!100.0!black}{-3.0} & \textcolor{red!100.0!black}{-2.1} & \textcolor{red!100.0!black}{-1.7} & \textcolor{red!100.0!black}{-3.0} & \textcolor{red!100.0!black}{-2.6} & \textcolor{red!100.0!black}{-1.1} & \textcolor{red!100.0!black}{-1.8} & \textcolor{red!100.0!black}{-1.3} & \textcolor{red!100.0!black}{-1.2} & \textcolor{red!100.0!black}{-1.2} \\
CorPipe individual & 71.1 & 76.3 & 78.7 & 76.9 & 65.7 & 62.0 & 73.8 & 63.2 & 79.5 & 66.8 & 64.8 & 73.4 & 71.7 & 69.9 \\
+ full mentions & \textcolor{blue!100.0!black}{+0.3} & \textcolor{red!100.0!black}{-1.9} & \textcolor{red!100.0!black}{-0.7} & \textcolor{blue!100.0!black}{+0.2} & \textcolor{blue!100.0!black}{+1.9} & \textcolor{blue!100.0!black}{+10.6} & \textcolor{red!100.0!black}{-0.7} & \textcolor{red!100.0!black}{-5.0} & \textcolor{red!100.0!black}{-1.3} & \textcolor{red!100.0!black}{-0.5} & \textcolor{red!100.0!black}{-1.9} & \textcolor{blue!100.0!black}{+1.3} & \textcolor{blue!100.0!black}{+2.3} & \textcolor{blue!100.0!black}{+0.3} \\
CorPipe best dev & @76.0 & @78.1 & @79.5 & @78.5 & @73.9 & @78.3 & @76.0 & @75.1 & @81.8 & @69.0 & @69.2 & @78.0 & @76.3 & @74.6 \\
+ full mentions & \textcolor{red!100.0!black}{-4.6} & \textcolor{red!100.0!black}{-3.6} & \textcolor{red!100.0!black}{-1.4} & \textcolor{red!100.0!black}{-1.3} & \textcolor{red!100.0!black}{-6.3} & \textcolor{red!100.0!black}{-5.6} & \textcolor{red!100.0!black}{-3.0} & \textcolor{red!100.0!black}{-16.9} & \textcolor{red!100.0!black}{-3.5} & \textcolor{red!100.0!black}{-2.6} & \textcolor{red!100.0!black}{-6.3} & \textcolor{red!100.0!black}{-3.3} & \textcolor{red!100.0!black}{-2.4} & \textcolor{red!100.0!black}{-4.4} \\
      \midrule
      \multicolumn{15}{c}{\textsc{B) Evaluation Including Singletons}} \\
CorPipe multilingual & 74.8 & 82.3 & 78.0 & 74.5 & 68.7 & 79.8 & 82.2 & 52.1 & 85.1 & 76.6 & 63.3 & 73.7 & 84.2 & 69.5 \\
CorPipe individual & 73.7 & 82.4 & 77.2 & 73.4 & 63.4 & 71.7 & 82.2 & 60.6 & 84.1 & 76.4 & 62.5 & 71.3 & 82.2 & 67.8 \\
CorPipe best dev & @77.5 & @83.2 & @78.0 & @74.7 & @70.4 & @82.6 & @83.2 & @71.6 & @85.8 & @77.5 & @66.8 & @74.8 & @84.7 & @73.0 \\
      \midrule
      \multicolumn{15}{c}{\textsc{C) Effect of Multilingual Data and the Pretrained Model}} \\
XLM-R base multilingual & 73.3 & 75.8 & 76.0 & 75.0 & 73.4 & 74.1 & 73.1 & @75.4 & 78.4 & 66.1 & 65.2 & @78.0 & 72.1 & 71.7 \\
XLM-R large multilingual & \textcolor{blue!76.0!black}{+1.5} & \textcolor{blue!100.0!black}{@+1.7} & \textcolor{blue!53.6!black}{+1.8} & \textcolor{blue!61.9!black}{+2.0} & \textcolor{blue!9.4!black}{+0.3} & \textcolor{blue!100.0!black}{@+4.1} & \textcolor{blue!89.1!black}{+2.1} & \textcolor{red!14.8!black}{-4.5} & \textcolor{blue!80.5!black}{+2.2} & \textcolor{blue!84.2!black}{+1.7} & \textcolor{blue!79.1!black}{+3.1} & \textcolor{red!0.2!black}{-0.0} & \textcolor{blue!77.1!black}{+2.9} & \textcolor{blue!29.9!black}{+0.9} \\
RemBERT multilingual & \textcolor{blue!100.0!black}{@+1.9} & \textcolor{blue!93.7!black}{+1.6} & \textcolor{blue!100.0!black}{@+3.3} & \textcolor{blue!100.0!black}{@+3.3} & \textcolor{blue!100.0!black}{@+2.9} & \textcolor{blue!58.5!black}{+2.4} & \textcolor{blue!100.0!black}{@+2.4} & \textcolor{red!19.9!black}{-6.1} & \textcolor{blue!100.0!black}{@+2.7} & \textcolor{blue!100.0!black}{@+2.0} & \textcolor{blue!100.0!black}{@+4.0} & \textcolor{red!13.2!black}{-1.2} & \textcolor{blue!100.0!black}{@+3.7} & \textcolor{blue!100.0!black}{@+2.9} \\
XLM-R base individual & \textcolor{red!99.6!black}{-4.6} & \textcolor{red!100.0!black}{-4.4} & \textcolor{red!100.0!black}{-0.3} & \textcolor{red!100.0!black}{-1.1} & \textcolor{red!33.8!black}{-7.8} & \textcolor{red!100.0!black}{-12.1} & \textcolor{red!100.0!black}{-1.9} & \textcolor{red!39.9!black}{-12.2} & \textcolor{red!100.0!black}{-2.8} & \textcolor{red!100.0!black}{-3.0} & \textcolor{red!100.0!black}{-3.8} & \textcolor{red!51.9!black}{-4.6} & \textcolor{red!100.0!black}{-2.3} & \textcolor{red!100.0!black}{-6.1} \\
XLM-R large individual & \textcolor{red!13.9!black}{-0.6} & \textcolor{blue!12.1!black}{+0.2} & \textcolor{blue!84.4!black}{+2.8} & \textcolor{blue!90.2!black}{+3.0} & \textcolor{red!33.5!black}{-7.7} & \textcolor{red!43.1!black}{-5.2} & \textcolor{red!46.6!black}{-0.9} & \textcolor{red!14.4!black}{-4.4} & \textcolor{blue!35.7!black}{+1.0} & \textcolor{blue!13.4!black}{+0.3} & \textcolor{blue!91.9!black}{+3.7} & \textcolor{red!60.5!black}{-5.4} & \textcolor{blue!95.1!black}{+3.5} & \textcolor{red!20.0!black}{-1.2} \\
RemBERT individual & \textcolor{red!100.0!black}{-4.7} & \textcolor{blue!32.8!black}{+0.6} & \textcolor{blue!83.8!black}{+2.8} & \textcolor{blue!58.5!black}{+1.9} & \textcolor{red!100.0!black}{-23.0} & \textcolor{red!99.9!black}{-12.1} & \textcolor{blue!27.6!black}{+0.7} & \textcolor{red!100.0!black}{-30.5} & \textcolor{blue!40.8!black}{+1.1} & \textcolor{blue!36.6!black}{+0.7} & \textcolor{red!11.2!black}{-0.4} & \textcolor{red!100.0!black}{-8.9} & \textcolor{blue!72.2!black}{+2.7} & \textcolor{red!29.7!black}{-1.8} \\
RemBERT 50\% additional & \textcolor{blue!17.7!black}{+0.3} & \textcolor{blue!55.2!black}{+1.0} & \textcolor{blue!74.0!black}{+2.5} & \textcolor{blue!73.2!black}{+2.4} & \textcolor{red!6.3!black}{-1.4} & \textcolor{red!4.5!black}{-0.5} & \textcolor{blue!72.4!black}{+1.7} & \textcolor{red!27.3!black}{-8.3} & \textcolor{blue!31.6!black}{+0.9} & \textcolor{blue!64.9!black}{+1.3} & \textcolor{blue!40.6!black}{+1.6} & \textcolor{red!39.0!black}{-3.5} & \textcolor{blue!98.1!black}{+3.6} & \textcolor{blue!58.5!black}{+1.7} \\
      \midrule
      \multicolumn{15}{c}{\textsc{D) Effect of Mixing Ratios using XLM-R base Pretrained Model}} \\
Logarithmic, w/o corpus id & @73.3 & 75.8 & 76.0 & 75.0 & @73.4 & @74.1 & @73.1 & @75.4 & 78.4 & 66.1 & 65.2 & 78.0 & 72.1 & @71.7 \\
Logarithmic, w/ corpus id & \textcolor{red!26.3!black}{-0.4} & \textcolor{red!42.0!black}{-0.5} & \textcolor{blue!7.4!black}{+0.1} & \textcolor{blue!24.5!black}{+0.3} & \textcolor{red!13.2!black}{-0.8} & \textcolor{red!8.7!black}{-0.3} & \textcolor{red!100.0!black}{-0.6} & \textcolor{red!66.0!black}{-4.6} & \textcolor{blue!100.0!black}{@+0.1} & \textcolor{blue!100.0!black}{@+0.6} & \textcolor{blue!100.0!black}{@+1.3} & \textcolor{red!78.2!black}{-0.9} & \textcolor{blue!100.0!black}{@+0.3} & \textcolor{red!41.9!black}{-0.7} \\
Uniform, w/o corpus id & \textcolor{red!51.9!black}{-0.8} & \textcolor{red!47.3!black}{-0.5} & \textcolor{red!28.8!black}{-0.2} & \textcolor{red!100.0!black}{-0.9} & \textcolor{red!28.3!black}{-1.8} & \textcolor{red!100.0!black}{-3.5} & \textcolor{red!26.6!black}{-0.2} & \textcolor{red!27.6!black}{-1.9} & \textcolor{blue!16.7!black}{+0.0} & \textcolor{red!100.0!black}{-0.0} & \textcolor{blue!33.3!black}{+0.4} & \textcolor{red!99.1!black}{-1.1} & \textcolor{blue!11.1!black}{+0.0} & \textcolor{red!91.3!black}{-1.5} \\
Uniform, w/ corpus id & \textcolor{red!100.0!black}{-1.6} & \textcolor{red!100.0!black}{-1.1} & \textcolor{red!100.0!black}{-0.5} & \textcolor{red!64.0!black}{-0.6} & \textcolor{red!100.0!black}{-6.4} & \textcolor{red!61.8!black}{-2.1} & \textcolor{red!68.7!black}{-0.4} & \textcolor{red!100.0!black}{-7.0} & \textcolor{blue!83.3!black}{+0.1} & \textcolor{blue!22.8!black}{+0.1} & \textcolor{red!100.0!black}{-0.5} & \textcolor{red!100.0!black}{-1.1} & \textcolor{red!100.0!black}{-0.6} & \textcolor{red!75.0!black}{-1.2} \\
Linear, w/o corpus id & \textcolor{red!16.7!black}{-0.3} & \textcolor{blue!100.0!black}{@+0.1} & \textcolor{blue!100.0!black}{@+0.8} & \textcolor{blue!100.0!black}{@+1.1} & \textcolor{red!17.5!black}{-1.1} & \textcolor{red!13.0!black}{-0.5} & \textcolor{red!75.0!black}{-0.5} & \textcolor{red!49.2!black}{-3.5} & \textcolor{red!100.0!black}{-0.1} & \textcolor{blue!59.6!black}{+0.3} & \textcolor{blue!72.0!black}{+1.0} & \textcolor{blue!100.0!black}{@+0.3} & \textcolor{red!17.5!black}{-0.1} & \textcolor{red!100.0!black}{-1.6} \\
      \midrule
      \multicolumn{15}{c}{\textsc{E) Effect of Mixing Ratios using RemBERT Pretrained Model}} \\
Logarithmic, w/o corpus id & 75.3 & 77.4 & 79.3 & 78.3 & 76.3 & 76.5 & 75.5 & 69.3 & 81.1 & 68.1 & @69.2 & 76.8 & 75.8 & @74.6 \\
Logarithmic, w/ corpus id & \textcolor{blue!100.0!black}{@+0.6} & \textcolor{blue!31.5!black}{+0.4} & \textcolor{blue!83.3!black}{+0.1} & \textcolor{blue!42.9!black}{+0.1} & \textcolor{blue!100.0!black}{@+3.0} & \textcolor{blue!100.0!black}{@+1.2} & \textcolor{red!6.7!black}{-0.1} & \textcolor{blue!100.0!black}{@+5.8} & \textcolor{blue!64.6!black}{+0.3} & \textcolor{blue!100.0!black}{@+0.9} & \textcolor{red!100.0!black}{-2.4} & \textcolor{red!100.0!black}{-1.3} & \textcolor{blue!84.6!black}{+0.1} & \textcolor{red!11.4!black}{-0.2} \\
Uniform, w/o corpus id & \textcolor{blue!21.4!black}{+0.1} & \textcolor{blue!98.4!black}{+1.2} & \textcolor{red!100.0!black}{-0.3} & \textcolor{red!50.0!black}{-0.1} & \textcolor{blue!81.7!black}{+2.4} & \textcolor{blue!40.2!black}{+0.5} & \textcolor{blue!100.0!black}{@+0.0} & \textcolor{red!100.0!black}{-0.9} & \textcolor{red!100.0!black}{-0.1} & \textcolor{blue!83.7!black}{+0.7} & \textcolor{red!27.5!black}{-0.6} & \textcolor{red!15.9!black}{-0.2} & \textcolor{blue!100.0!black}{@+0.1} & \textcolor{red!87.1!black}{-1.2} \\
Uniform, w/ corpus id & \textcolor{red!91.7!black}{-0.1} & \textcolor{red!100.0!black}{-0.0} & \textcolor{red!48.4!black}{-0.2} & \textcolor{red!100.0!black}{-0.3} & \textcolor{red!100.0!black}{-4.2} & \textcolor{blue!21.3!black}{+0.3} & \textcolor{red!7.3!black}{-0.1} & \textcolor{blue!77.4!black}{+4.5} & \textcolor{blue!77.1!black}{+0.4} & \textcolor{blue!67.4!black}{+0.6} & \textcolor{red!40.7!black}{-1.0} & \textcolor{red!9.8!black}{-0.1} & \textcolor{blue!46.2!black}{+0.1} & \textcolor{red!91.7!black}{-1.2} \\
Linear, w/o corpus id & \textcolor{red!100.0!black}{-0.1} & \textcolor{blue!100.0!black}{@+1.3} & \textcolor{blue!100.0!black}{@+0.1} & \textcolor{blue!100.0!black}{@+0.2} & \textcolor{red!54.2!black}{-2.3} & \textcolor{red!100.0!black}{-0.5} & \textcolor{red!100.0!black}{-1.5} & \textcolor{blue!32.8!black}{+1.9} & \textcolor{blue!100.0!black}{@+0.5} & \textcolor{blue!76.7!black}{+0.7} & \textcolor{red!43.6!black}{-1.0} & \textcolor{blue!100.0!black}{@+0.4} & \textcolor{blue!7.7!black}{+0.0} & \textcolor{red!100.0!black}{-1.3} \\
      \midrule
      \multicolumn{15}{c}{\textsc{F) Zero-shot Evaluation of a Multilingual Model}} \\
Multilingual XLM-R base & 73.3 & 75.8 & 76.0 & 75.0 & 73.4 & 74.1 & 73.1 & @75.4 & 78.4 & 66.1 & 65.2 & @78.0 & 72.1 & 71.7 \\
Zero-shot XLM-R base & \textcolor{red!100.0!black}{-17.1} & \textcolor{red!100.0!black}{-11.1} & \textcolor{red!100.0!black}{-28.6} & \textcolor{red!100.0!black}{-23.8} & \textcolor{red!100.0!black}{-13.3} & \textcolor{red!100.0!black}{-13.8} & \textcolor{red!100.0!black}{-19.8} & \textcolor{red!100.0!black}{-18.5} & \textcolor{red!100.0!black}{-6.8} & \textcolor{red!91.0!black}{-7.6} & \textcolor{red!100.0!black}{-16.1} & \textcolor{red!100.0!black}{-23.8} & \textcolor{red!100.0!black}{-24.6} & \textcolor{red!100.0!black}{-15.1} \\
Multilingual RemBERT & \textcolor{blue!100.0!black}{@+1.9} & \textcolor{blue!100.0!black}{@+1.6} & \textcolor{blue!100.0!black}{@+3.3} & \textcolor{blue!100.0!black}{@+3.3} & \textcolor{blue!100.0!black}{@+2.9} & \textcolor{blue!100.0!black}{@+2.4} & \textcolor{blue!100.0!black}{@+2.4} & \textcolor{red!32.9!black}{-6.1} & \textcolor{blue!100.0!black}{@+2.7} & \textcolor{blue!100.0!black}{@+2.0} & \textcolor{blue!100.0!black}{@+4.0} & \textcolor{red!5.0!black}{-1.2} & \textcolor{blue!100.0!black}{@+3.7} & \textcolor{blue!100.0!black}{@+2.9} \\
Zero-shot RemBERT & \textcolor{red!72.9!black}{-12.5} & \textcolor{red!60.1!black}{-6.7} & \textcolor{red!82.8!black}{-23.7} & \textcolor{red!86.4!black}{-20.6} & \textcolor{red!83.5!black}{-11.1} & \textcolor{red!54.2!black}{-7.5} & \textcolor{red!78.7!black}{-15.6} & \textcolor{red!53.0!black}{-9.8} & \textcolor{red!42.0!black}{-2.8} & \textcolor{red!100.0!black}{-8.3} & \textcolor{red!65.5!black}{-10.5} & \textcolor{red!84.0!black}{-20.0} & \textcolor{red!74.5!black}{-18.3} & \textcolor{red!47.8!black}{-7.2} \\
      \midrule
      \multicolumn{15}{c}{\textsc{G) Effect of Several Language-specific base Pretrained Models}} \\
XLM-R base individual & 68.7 & 71.4 & 75.7 & 73.9 & 65.7 & 62.0 & 71.2 & 63.2 & 75.6 & 63.1 & 61.5 & 73.4 & 69.8 & 65.6 \\
mBERT \rlap{\citep{devlin-etal-2019-bert}} & \textcolor{red!100.0!black}{-2.8} & \textcolor{red!100.0!black}{-1.5} & \textcolor{red!100.0!black}{-3.0} & \textcolor{red!100.0!black}{-3.4} & \textcolor{red!22.0!black}{-3.3} & \textcolor{blue!2.3!black}{+0.4} & \textcolor{red!100.0!black}{-2.8} & \textcolor{red!6.2!black}{-1.1} & \textcolor{red!100.0!black}{-1.8} & \textcolor{red!100.0!black}{-1.1} & \textcolor{red!100.0!black}{-2.7} & \textcolor{red!100.0!black}{-7.5} & \textcolor{red!100.0!black}{-4.4} & \textcolor{red!100.0!black}{-3.6} \\
BERTa \rlap{\scriptsize\citep{armengol-estape-etal-2021-multilingual}} &  & \textcolor{blue!21.6!black}{+1.3} &  &  &  &  &  &  &  &  &  &  &  &  \\
RobeCzech \rlap{\citep{RobeCzech2021}} &  &  & \textcolor{blue!55.0!black}{+2.0} & \textcolor{blue!62.7!black}{+2.8} &  &  &  &  &  &  &  &  &  &  \\
gBERT \rlap{\citep{chan-etal-2020-germans}} &  &  &  &  & \textcolor{red!65.4!black}{-9.9} & \textcolor{blue!32.7!black}{+5.3} &  &  &  &  &  &  &  &  \\
SpanBERT \rlap{\citep{joshi-etal-2020-spanbert}} &  &  &  &  &  &  & \textcolor{red!13.1!black}{-0.4} & \textcolor{red!13.3!black}{-2.4} &  &  &  &  &  &  \\
BETO \rlap{\citep{Beto2020}} &  &  &  &  &  &  &  &  & \textcolor{blue!7.8!black}{+0.4} &  &  &  &  &  \\
CamemBERT \rlap{\citep{martin-etal-2020-camembert}} &  &  &  &  &  &  &  &  &  & \textcolor{red!20.7!black}{-0.2} &  &  &  &  \\
HuBERT \rlap{\citep{huBERT2020}} &  &  &  &  &  &  &  &  &  &  & \textcolor{blue!47.1!black}{+3.6} &  &  &  \\
LitLatBERT \rlap{\citep{LitLatBERT21}} &  &  &  &  &  &  &  &  &  &  &  & \textcolor{blue!59.3!black}{+2.7} &  &  \\
HerBERT \rlap{\citep{mroczkowski-etal-2021-herbert}} &  &  &  &  &  &  &  &  &  &  &  &  & \textcolor{blue!27.3!black}{+1.6} &  \\
RuBERT \rlap{\citep{DeepPavlov2019}} &  &  &  &  &  &  &  &  &  &  &  &  &  & \textcolor{blue!2.3!black}{+0.2} \\
XLM-R large individual & \textcolor{blue!60.9!black}{+4.0} & \textcolor{blue!74.9!black}{+4.6} & \textcolor{blue!85.6!black}{+3.1} & \textcolor{blue!92.8!black}{+4.1} & \textcolor{blue!0.5!black}{+0.0} & \textcolor{blue!42.5!black}{+6.9} & \textcolor{blue!23.8!black}{+1.0} & \textcolor{blue!100.0!black}{@+7.8} & \textcolor{blue!68.2!black}{+3.8} & \textcolor{blue!65.5!black}{+3.3} & \textcolor{blue!95.9!black}{+7.4} & \textcolor{red!10.2!black}{-0.8} & \textcolor{blue!97.0!black}{+5.8} & \textcolor{blue!53.9!black}{+4.8} \\
RemBERT individual & \textcolor{red!0.7!black}{-0.0} & \textcolor{blue!80.8!black}{+4.9} & \textcolor{blue!85.1!black}{+3.1} & \textcolor{blue!69.2!black}{+3.1} & \textcolor{red!100.0!black}{-15.2} & \textcolor{blue!0.1!black}{+0.0} & \textcolor{blue!60.0!black}{+2.6} & \textcolor{red!100.0!black}{-18.3} & \textcolor{blue!70.8!black}{+3.9} & \textcolor{blue!74.8!black}{+3.8} & \textcolor{blue!43.2!black}{+3.3} & \textcolor{red!57.0!black}{-4.3} & \textcolor{blue!82.8!black}{+5.0} & \textcolor{blue!47.3!black}{+4.3} \\
XLM-R large multilingual & \textcolor{blue!93.0!black}{+6.1} & \textcolor{blue!100.0!black}{@+6.1} & \textcolor{blue!57.2!black}{+2.1} & \textcolor{blue!71.7!black}{+3.2} & \textcolor{blue!75.5!black}{+8.0} & \textcolor{blue!100.0!black}{@+16.2} & \textcolor{blue!94.0!black}{+4.1} & \textcolor{blue!98.6!black}{+7.7} & \textcolor{blue!90.4!black}{+5.0} & \textcolor{blue!93.7!black}{+4.8} & \textcolor{blue!89.3!black}{+6.9} & \textcolor{blue!100.0!black}{@+4.6} & \textcolor{blue!85.8!black}{+5.1} & \textcolor{blue!77.1!black}{+6.9} \\
RemBERT multilingual & \textcolor{blue!100.0!black}{@+6.6} & \textcolor{blue!98.2!black}{+6.0} & \textcolor{blue!100.0!black}{@+3.6} & \textcolor{blue!100.0!black}{@+4.4} & \textcolor{blue!100.0!black}{@+10.6} & \textcolor{blue!89.5!black}{+14.5} & \textcolor{blue!100.0!black}{@+4.3} & \textcolor{blue!78.6!black}{+6.1} & \textcolor{blue!100.0!black}{@+5.5} & \textcolor{blue!100.0!black}{@+5.1} & \textcolor{blue!100.0!black}{@+7.7} & \textcolor{blue!74.9!black}{+3.5} & \textcolor{blue!100.0!black}{@+6.0} & \textcolor{blue!100.0!black}{@+9.0} \\
      \midrule
      \multicolumn{15}{c}{\textsc{H) Effect of the Right Context Size; Development Version}} \\
Right context 0 & 67.4 & @70.7 & 75.0 & 73.6 & 59.2 & 62.4 & @68.3 & 68.6 & @74.4 & 61.1 & 59.2 & 71.5 & 69.2 & 62.2 \\
Right context 50 & \textcolor{blue!100.0!black}{@+0.8} & \textcolor{red!34.7!black}{-0.4} & \textcolor{blue!83.8!black}{+1.3} & \textcolor{blue!85.9!black}{+0.6} & \textcolor{blue!76.4!black}{+3.9} & \textcolor{blue!65.1!black}{+1.8} & \textcolor{red!100.0!black}{-0.7} & \textcolor{blue!100.0!black}{@+1.2} & \textcolor{red!95.8!black}{-0.5} & \textcolor{blue!6.9!black}{+0.0} & \textcolor{blue!80.0!black}{+1.7} & \textcolor{blue!100.0!black}{@+0.5} & \textcolor{blue!50.0!black}{+0.3} & \textcolor{blue!100.0!black}{@+1.7} \\
Right context 100 & \textcolor{blue!73.8!black}{+0.6} & \textcolor{red!100.0!black}{-1.2} & \textcolor{blue!100.0!black}{@+1.5} & \textcolor{blue!100.0!black}{@+0.7} & \textcolor{blue!100.0!black}{@+5.1} & \textcolor{blue!100.0!black}{@+2.7} & \textcolor{red!9.6!black}{-0.1} & \textcolor{red!100.0!black}{-3.8} & \textcolor{red!100.0!black}{-0.5} & \textcolor{blue!100.0!black}{@+0.6} & \textcolor{blue!100.0!black}{@+2.1} & \textcolor{red!100.0!black}{-0.2} & \textcolor{blue!100.0!black}{@+0.6} & \textcolor{blue!39.3!black}{+0.7} \\
      \midrule
      \multicolumn{15}{c}{\textsc{I) Effect of the Maximum Number of Links during Training; Development Version}} \\
Unlimited & @67.4 & @70.7 & 75.0 & @73.6 & 59.2 & @62.4 & @68.3 & @68.6 & @74.4 & @61.1 & @59.2 & 71.5 & @69.2 & @62.2 \\
At most 1 link & \textcolor{red!100.0!black}{-3.8} & \textcolor{red!100.0!black}{-3.3} & \textcolor{red!100.0!black}{-0.9} & \textcolor{red!100.0!black}{-3.6} & \textcolor{red!100.0!black}{-3.1} & \textcolor{red!100.0!black}{-4.3} & \textcolor{red!100.0!black}{-4.8} & \textcolor{red!100.0!black}{-8.7} & \textcolor{red!100.0!black}{-4.0} & \textcolor{red!100.0!black}{-3.1} & \textcolor{red!100.0!black}{-2.6} & \textcolor{red!100.0!black}{-5.0} & \textcolor{red!100.0!black}{-3.0} & \textcolor{red!100.0!black}{-3.8} \\
At most 2 links & \textcolor{red!35.5!black}{-1.4} & \textcolor{red!42.8!black}{-1.4} & \textcolor{blue!35.8!black}{+0.2} & \textcolor{red!55.2!black}{-2.0} & \textcolor{blue!44.0!black}{+1.5} & \textcolor{red!56.9!black}{-2.4} & \textcolor{red!38.7!black}{-1.9} & \textcolor{red!63.6!black}{-5.6} & \textcolor{red!19.4!black}{-0.8} & \textcolor{red!3.2!black}{-0.1} & \textcolor{red!38.3!black}{-1.0} & \textcolor{red!10.1!black}{-0.5} & \textcolor{red!89.2!black}{-2.6} & \textcolor{red!33.5!black}{-1.3} \\
At most 3 links & \textcolor{red!15.1!black}{-0.6} & \textcolor{red!27.2!black}{-0.9} & \textcolor{blue!100.0!black}{@+0.5} & \textcolor{red!6.1!black}{-0.2} & \textcolor{blue!100.0!black}{@+3.5} & \textcolor{red!9.6!black}{-0.4} & \textcolor{red!7.2!black}{-0.4} & \textcolor{red!74.7!black}{-6.5} & \textcolor{red!29.4!black}{-1.2} & \textcolor{red!11.7!black}{-0.4} & \textcolor{red!7.2!black}{-0.2} & \textcolor{blue!100.0!black}{@+1.1} & \textcolor{red!63.7!black}{-1.9} & \textcolor{red!15.4!black}{-0.6} \\
      \bottomrule
    \end{tabular}

    \caption{Ablation experiments evaluated on the development sets (CoNLL score in \%). In A) and B), the scores of the official submissions are used; in C) to I), we report the highest development set score from any epoch.}
    \label{tab:dev_ablations}
\end{table*}

Official results of the \CRAC~on the test set can be found in Table~\ref{tab:official_results}. Our multilingual models, \textbf{\itshape{best dev}} and \textbf{\itshape{multilingual}}, scored 1st and 2nd, respectively, while our \textbf{\itshape{individual}} models trained on each dataset placed 4th. 

\subsection{Multilingual Models}
\label{sec:results-multilingual}

A view on the effectiveness of multilingual models is shown in official ablation results on test data in Table~\ref{tab:official_ablations}, which compares all our three individual/multilingual settings: \textbf{\itshape{multilingual}} as a baseline, \textbf{\itshape{individual}} and \textbf{{\itshape{best dev}}}, using a base encoder (XLM-R base for the multilingual baseline, best-performing base encoder for the remaining cases) and a large encoder (RemBERT). The \textbf{\itshape{multilingual}} is superior to \textbf{\itshape{individual}} for all datasets, with the exception of the three largest datasets using a base encoder
-- we hypothesize that the base encoder does not have sufficient capacity to capture the largest datasets in the \textbf{\itshape{multilingual}} setting, because with a large encoder, also the three largest datasets benefit from the \textbf{\itshape{multilingual}} model. %
Furthermore, Table~\ref{tab:dev_ablations}.C demonstrates that while XLM-R large is the best in the \textbf{\itshape{individual}} settings, RemBERT delivers superior \textbf{\itshape{multilingual}} performance.

Motivated by the improvements of the multilingual models, we considered a setting where 50\% of the training data comes from a single dataset and the rest from all other datasets (with logarithmic mixing). Surprisingly, such setting delivers consistently worse performance than the multilingual models (last line of Table~\ref{tab:dev_ablations}.C).

The comparison of \textbf{\itshape{logarithmic}}, \textbf{\itshape{uniform}}, and \textbf{\itshape{linear}} mixing, together with the presence or absence of \textbf{\itshape{corpus id}}, is evaluated in Table~\ref{tab:dev_ablations}.D and Table~\ref{tab:dev_ablations}.E. Unexpectedly, neither the mixing rations nor the \textbf{\textit{corpus id}} have a large effect on the results, which is surprising especially for the \textbf{\textit{linear}} mixing, where the smallest treebanks are nearly 100 times less frequent than the largest one.

\subsection{Zero-shot Evaluation}
The prospect of not including the corpus id opens an interesting possibility of using the model in zero-shot setting, i.e., on a different language than it was trained on. To perform such zero-shot evaluation, we trained for every language a multilingual model \textit{without} datasets in this language, and then evaluated the model on them. The results, presented in Table~\ref{tab:dev_ablations}.F, were below our expectations, slightly surpassing 60\% macro average on the development set with the RemBERT model.

\subsection{Monolingual Pretrained Language Models}

Table~\ref{tab:dev_ablations}.G presents the evaluation of the best-performing monolingual base-sized pretrained models we found. While the specialized models consistently surpass mBERT and are mostly better than XLM-R base, they are all worse than the individual XLM-R large models (with the exception of Lithuanian) and even more dominated by the RemBERT multilingual model. This indicates that, nowadays, pretraining a base-sized monolingual BERT model has merit only in improving the running time, not model performance, when large pretrained multilingual models are now available.

\subsection{Context Window}
\label{sec:results-right_context}

Table~\ref{tab:dev_ablations}.H shows the effect of using a right context of size 0, 50, and 100. The evaluation, performed on a base-sized model with a preliminary, development version of CorPipe, shows that the presence of the right context is beneficial, but does not clearly indicate whether context of size 100 is better than 50.

\subsection{Number of Links}
\label{sec:results-links}

The effect of limiting the number of predicted antecedents during training is presented by Table~\ref{tab:dev_ablations}.I. The evaluation (performed again on a base-sized model with a preliminary, development version of CorPipe) shows that performance increases with the number of antecedents considered during training.

\subsection{Singletons}
\label{sec:results-singletons}

Singletons (entities with only one mention in the document) were excluded from the official evaluation primary metric. Our antecedent-maximization strategy however accounts for them by adding ``self'' to antecedent candidates pool. We publish the official evaluation with singletons on the test set in Table~\ref{tab:official_singletons} and the ablation evaluation with singletons on the dev set in Table~\ref{tab:dev_ablations}.B.

\subsection{Limiting Mention Spans to Their Heads}
\label{sec:results-heads}

Comparison between full predicted mention spans and the predicted spans reduced to their syntactic heads in Table~\ref{tab:dev_ablations}.A shows that \textit{partial matching} favors post-processing which keeps syntactic heads and avoids ``overprediction'' beyond the gold mention span.

\section{Conclusions}

We presented a jointly trained pipeline approach as a winning contribution to the \CRAClong~\cite{sharedtask-findings}. We published a thorough comparison of pretrained large language models for the task. Finally, we focused on multilingual models and we conclude that one multilingual, all-data model with large encoder outperformed individual monolingual fine-tuned models on all datasets. The source code is available at {\small\url{https://github.com/ufal/crac2022-corpipe}}.

\section*{Acknowledgements}

This work has been supported by the Grant Agency of the Czech Republic, project EXPRO LUSyD (GX20-16819X), and has been using data provided by the LINDAT/CLARIAH-CZ Research Infrastructure ({\footnotesize\url{https://lindat.cz}}) of the Ministry of Education, Youth and Sports of the Czech Republic (Project No. LM2018101).

\bibliography{anthology,custom}
\bibliographystyle{acl_natbib}

\appendix

\end{document}